\documentclass[tinyml]{acmart}

\AtBeginDocument{%
  \providecommand\BibTeX{{%
    \normalfont B\kern-0.5em{\scshape i\kern-0.25em b}\kern-0.8em\TeX}}}

\setcopyright{rightsretained}
\copyrightyear{2021}
\acmYear{2021}

\usepackage{mathtools}
\usepackage{graphicx}
\usepackage{fixltx2e}
\usepackage{url}
\usepackage{siunitx}
\usepackage{booktabs}
\usepackage{multirow}
\usepackage{libertine}
 \usepackage{color}
\usepackage{array}
\begin{document}

\title{Characterization of Neural Networks Automatically \\Mapped on Automotive-grade Microcontrollers}

\author{Giulia Crocioni}
\email{giulia.crocioni@mail.polimi.it}
\author{Giambattista Gruosso}
\orcid{0000-0001-6417-3750}
\email{giambattista.gruosso@polimi.it}
\affiliation{%
  \institution{Politecnico di Milano}
\institution{DEIB}
  \streetaddress{Piazza Leonardo da Vinci, 32}
  \city{Milano}
  \country{Italy}
  \postcode{I-20133}
}

\author{Danilo Pau}
\email{danilo.pau@st.com}
\author{Davide Denaro}
\email{davide.denaro@st.com}
\affiliation{%
  \institution{STMicroelectronics}
  \streetaddress{Via C. Olivetti, 2}
  \city{Agrate Brianza}
  \country{Italy}
    \postcode{I-20864}
  }

\author{Luigi Zambrano}
\email{luigi.zambrano@st.com}
\author{Giuseppe Di Giore}
\email{giuseppe.di-giore@st.com}
\affiliation{%
  \institution{STMicroelectronics}
  \city{Arzano (NA)}
  \country{Italy}
  }

\renewcommand{\shortauthors}{Crocioni Gruosso, et al.}

\begin{abstract}
Nowadays, Neural Networks represent a major expectation for the realization of powerful Deep Learning algorithms, which are able to determine the behaviors and operations of several physical systems. Computational resources required for model, training and running are large, especially when related to the amount of data that Neural Networks typically need to be able to generalize. The latest TinyML technologies allow the integration of pre-trained models on embedded systems, giving the opportunity to make computing at the edge faster, cheaper, and safer. Although these technologies originated in the consumer and industrial worlds, there are many sectors that can greatly benefit from them, such as the automotive industry. In this paper, we present a framework for implementing Neural Network-based models on a family of automotive Microcontrollers, showing their efficiency in two case studies applied to vehicles: intrusion detection on the Controller Area Network bus, and residual capacity estimation in Lithium-Ion batteries, widely used in Electric Vehicles.
\end{abstract}

\keywords{datasets, neural networks, gaze detection, text tagging}

\maketitle
\section{Introduction}
Machine Learning (ML) is one of the most groundbreaking technologies driving the fourth industrial revolution, enabling the mining of valuable information from large amounts of data that the human brain alone is not able to capture \cite{SYAM2018135, 9166461}. In recent years, research in ML is evolving at a huge speed, bringing great advances in several ML sub-fields, such as Deep Learning (DL), Deep Neural Networks (DNNs), and Deep Generative Models (DGMs) \cite{deep_l_overview}. Primary sources that generate data include social media platforms, industrial control systems, and sensors used in the Internet of Things (IoT). Developing meaningful and reliable models using such a large amount of information requires extremely powerful resources in terms of processing capabilities. However, in recent years there has been a growing interest in embedding ML models into Microcontroller Units (MCUs), small computers with integrated memory which are often the core of embedded technology \cite{warden_tinyml_2020, david2020tensorflow, 9166461}. Embedded systems are widely used in healthcare, energy, consumer electronics, and automotive, and are often required to process information and make real-time decisions relying on resource constrained hardware. Thanks to the latest TinyML technologies, pre-trained models can be integrated on embedded systems, giving the opportunity to make computing at the edge faster, cheaper, and safer.

Although TinyML technologies originated in the consumer and industrial worlds, there are many sectors that can greatly benefit from them, such as the automotive industry \cite{a1,a2,a3,a4}. In fact, automotive control units are typically limited in terms of computational power and data storage capacity.
The goal of architectures embedded on automotive MCUs is to analyze data as they are received, while the training phase is performed offline on more powerful systems. Therefore, in the tool-chain it is essential to adopt efficient models whose training phase takes place offline, and to consider that the on-board control units must perform the primary functions for which they are designed, from the on-board devices management to the engine control.

In this paper, we present a framework for implementing NN-based models on a family of automotive MCUs, showing their efficiency in two case studies applied to vehicles: intrusion detection on the Controller Area Network (CAN) bus, and residual capacity estimation in Lithium-Ion (Li-Ion) batteries, widely used in Electric Vehicles (EVs). The rest of the paper is organized as follows. In Section \ref{sec:rel_works} related works on TinyML approaches and automotive applications are reviewed. Section \ref{sec:1} describes the family of automotive MCUs chosen for the experiments, and the development tool used for models integration on the MCUs. The following Sections present two TinyML case studies applied to vehicles. In particular, Section \ref{sec:cas1} describes an Intrusion Detection System (IDS) on CAN bus traffic, and Section \ref{sec:cas2} describes an architecture for the estimation of the remaining releasable capacity of Li-Ion batteries. Section \ref{sec:compl} presents an accurate and original analysis of the models complexity. Finally, conclusions are summarized in Section \ref{sec:con}, in which the main findings of the paper are discussed.

\section{Related Works}\label{sec:rel_works}

In recent years, modern vehicles have been equipped with dozens of Electronic Control Units (ECUs), embedded systems that control many vehicles functions, and whose functional requirements have become increasingly complex \cite{daimi_securing_2016}.

\begin{figure}[h!]
  \caption{In-vehicle communication system example, taken from \cite{HITBSecConf}.}
  \centering
    \includegraphics[width=0.9\columnwidth]{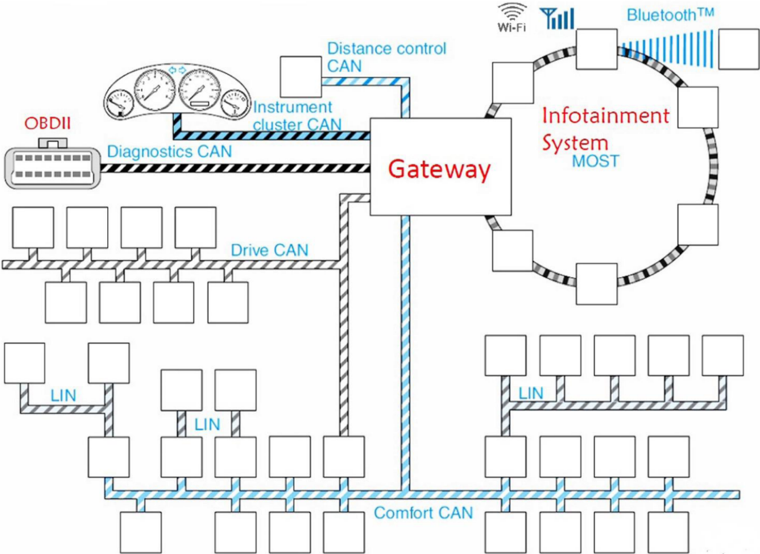}
    \label{fig:commsystem}
\end{figure}

Figure \ref{fig:commsystem} shows an example of an In-vehicle communication system \cite{HITBSecConf} that contains several sub-networks with different functionalities interconnected by means of a gateway. 
In such a system, it becomes increasingly important to use traditional control techniques combined with ML and Artificial Intelligence (AI) methods. Since MCUs in automotive embedded electronics are resource-constrained, there is a growing interest in integrating TinyML technologies, that have already been widely used in other consumer applications.
New ML solutions in automotive are being developed each year in both industry and literature \cite{a1,a2,a3,a4}. In this paper, we present a methodology to embed ML models into automotive-grade MCUs, and we apply it on two typical applications, that are described below.

\subsection{Intrusion Detection Systems}

A CAN bus is a multi-master message protocol, which is used to interconnect ECUs in automotive systems (e.g., modern vehicles) \cite{canbosh}. This communication protocol meets the requirements for real-time communication, and low implementation cost. With a maximum transfer rate of 1 Mbit/sec, CAN bus is the de facto standard for ECUs interconnections \cite{HITBSecConf}. Vehicles manufacturers must carefully design interconnections between critical and non-critical sub-networks, trying to prevent attacks on ECUs, such as the ones that exploit vulnerabilities in the infotainment system. Researchers demonstrated that it is possible to exploit a feature defined in the CAN standard, and to disable the braking system \cite{wired, remoteex}. Although the CAN protocol was originally designed to be secure, every vehicle using CAN bus is open to attack \cite{lokman_intrusion_2019}. Some CAN bus vulnerabilities already found by researchers are listed below:

\begin{itemize}
    \item It is a multicast message protocol with no intrinsic addressing and authentication mechanism. Therefore, a compromised ECU can access every message in its sub-network, and can send messages with a false identity.
    \item It is a bandwidth-constrained protocol for nowadays vehicles, making it challenging to introduce message encryption.
    \item Most nodes are automotive-grade MCUs with limited memory and computational capability, making it difficult to implement complex security protocols.
\end{itemize}

The introduction of an IDS can be a suitable countermeasure to CAN bus vulnerabilities.
One of the intrusion detection methods is the anomaly-based approach.
An intuitive description of this method is to consider a monitoring system, an ECU, that listens to CAN bus traffic and learns normal behavior. The intruder activities raise abnormal traffic, and this alerts the trained IDS.

\subsection{Battery Releasable Capacity}
In recent years, Li-Ion batteries are receiving significant interest because of their several advantages in terms of high specific energy and power \cite{marano_lithium-ion_2009}. Rechargeable battery stacks based on Li-Ion cells are used to power many systems, including portable devices, such as smartphones, and automotive systems, such as Hybrid Electric Vehicles (HEV) and Electric Vehicles (EV) \cite{applic_batt, 8605007, 8493203}. 
To increase the safety, reliability, and cost-effectiveness of a battery, it is essential to improve the Battery Management System (BMS) features. \cite{BMS}. In this regard, battery capacity estimation is essential as it allows the calculation of the State of Health (SoH). SoH is a measure of battery functionality in energy storage and delivery, and represents a fundamental parameter for the Battery Health Monitoring (BHM) \cite{9133084}. Due to internal aging processes, capacity decays over the battery's lifetime even if it is not used, causing battery performance to decrease. Typically, a 20\% reduction in rated capacity is considered the limit for safe use of the component (i.e., ${C_{max}}\leq0.8 C_{rated}$), under which the battery performance may not be reliable \cite{choi_machine_2019, nxp_tja1043}. Therefore, SoH diagnosis and accurate releasable capacity estimation are essential for safety risks reduction, critical failure prevention, and appropriate battery replacement \cite{capacity}. Data-driven methods based on ML techniques are widely used for battery capacity estimation \cite{choi_machine_2019}. They compute the releasable capacity starting from measurable parameters, such as voltage and current, which can be easily extracted from a vehicle via CAN bus \cite{minxin_zheng_li-ion_2008}. Since SoH is highly non-linear and not directly observable, DL algorithms are shown to be more flexible and efficient than traditional methods \cite{chemali_intelligent_nodate}.

\section{SPC5-Studio-AI: automated conversion of pre-trained Neural Networks \label{sec:1}}


SPC5-Studio-AI is a plug-in component of the SPC5-STUDIO development environment, which supports the Chorus SPC58 MCU family. This tool provides the capability to automatically generate, execute, and validate pre-trained NN models on automotive-grade MCUs. It converts the models to optimized ANSI C code for SPC58 MCUs, and automatically profiles their complexity. Several DL frameworks are supported, such as Keras, TensorFlow Lite, ONNX, Lasagne, Caffe, and ConvNetJS. 
Thanks to a well-defined short number of public APIs, the ANSI C library can be imported into application-specific projects, such as the two case studies presented in Sections \ref{sec:cas1} and \ref{sec:cas2}. Moreover, SPC5-Studio-AI provides NN validation and complexity profiling measuring key metrics such as validation error, memory requirements (i.e. Flash and RAM), and execution
time, directly on MCUs. This plugin is integrated within SPC5-STUDIO development environment (currently version 6.0.0). 

\begin{figure}[h!]
  \caption{SPC5-Studio.AI block diagram} 
  \centering
    \includegraphics[width=0.9\columnwidth]{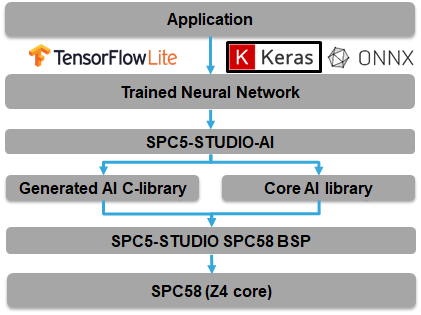}
    \label{fig:can_struct1}
\end{figure}

SPC5-Studio.AI was used to embed and to validate the developed NNs on three automotive-grade MCUs, suitable for applications which require low-power, connectivity and security \cite{spc5}: SPC584B, SPC58EC, and SPC58NH. The main features of the chosen MCUs are shown in Table \ref{table:mcus}. Power consumption was computed considering each MCU at its maximum frequency, and with all cores enabled. 

\begin{table}[h!]
\centering
\caption{Main features of the automotive-grade MCUs used for the complexity analysis.}
\label{table:mcus} 
\resizebox{0.47\textwidth}{!}{
\begin{tabular}{@{}ccccccc@{}}
\toprule
\textbf{Device} &
  \textbf{\begin{tabular}[c]{@{}c@{}}Flash\\ {[}Mb{]}\end{tabular}} &
  \textbf{\begin{tabular}[c]{@{}c@{}}RAM\\ {[}Mb{]}\end{tabular}} &
  \textbf{\begin{tabular}[c]{@{}c@{}}Clock\\ {[}MHz{]}\end{tabular}} &
  \textbf{I/D Cache} &
  \textbf{FPU} &
  \textbf{\begin{tabular}[c]{@{}c@{}}Power\\ Consumption\\ {[}mA{]}\end{tabular}} \\ \midrule
SPC584B & 2  & 192  & 120 & Yes & Yes & 102.0 \\
SPC58EC & 4  & 512  & 180 & Yes & Yes & 132.6 \\
SPC58NH & 10 & 1024 & 200 & Yes & Yes & 239.6 \\ \bottomrule
\end{tabular}}
\end{table}

\section{Case Study 1: Intrusion detection in an automotive network \label{sec:cas1}}

In this Section, we present a Long Short-Term Memory (LSTM) Autoencoder to detect CAN bus anomalies raised by abnormal traffic, using the SynCAN dataset \cite{9044377}. After embedding and validating the pre-trained Autoencoder into the aforementioned automotive MCUs, its complexity was profiled. 

\subsection{Dataset}

\begin{figure}[h!]
  \caption{The structure of a CAN data frame, taken from \cite{Cho2016FingerprintingEC}.}
  \centering
    \includegraphics[width=0.9\columnwidth]{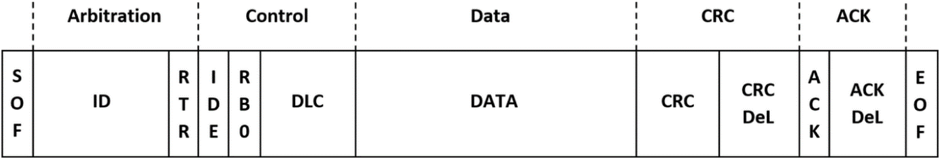}
    \label{fig:can_struct}
\end{figure}

A CAN bus is a multi-master message broadcast system, in which each message (or packet) has a standard format consisting of a timestamp, an identifier ID, and up to 8 bytes of payload. The ID represents the type of message, and the payload may carry one or more meaningful signals. Thus, the CAN bus traffic in a subnetwork can be represented as time series. Figure \ref{fig:can_struct} shows the typical structure of a CAN data frame as referenced in  \cite{Cho2016FingerprintingEC}.


The data for the analysis were taken from SynCAN (Synthetic CAN bus data), a synthetic dataset created to benchmark, evaluate, and compare different CAN IDSs on different attack scenarios \cite{9044377}.  The dataset is composed of normal and abnormal traffic signals, the latter divided according to the attack type:

\begin{enumerate}
	\item Plateau attack. A signal overwritten to a constant value over a time period.
	\item Continuous change attack. A  signal slowly drifted from its true value.
	\item Playback attack. An already recorded time series of values of the signal itself, over a time period.
	\item Suppress attack. A signal completely suppressed.
	\item Flooding attack. A signal to deny access to the other ECUs.
\end{enumerate}

The anomaly detector developed was trained on the normal traffic signals, and was tested on the abnormal ones, corresponding to the attacks.

\subsection{Long Short-Term Memory Autoencoder}

The architecture used for the anomaly detector was an LSTM Autoencoder, that demonstrated to be effective in learning the normal behavior of a simulated CAN bus traffic, as shown by \cite{DBLP:journals/corr/abs-1906-02492}. 

The implemented Autoencoder consists of a dense layer, two LSTM layers, and a dense output layer. Input features are 24 consecutive messages related to network traffic of 20 different signals. Thus, input data are provided in the three-dimensional format: number of samples, time steps (24), and features (20). The LSTM layers have 18 output units, and the dense output layer consists of 20 nodes. The network is made by 6272 parameters. The overall topology is shown in Table \ref{tab:topology}.

\begin{table}[h!]
\centering
\caption{The implemented Autoencoder consists of a dense layer, two LSTM layers, and a dense output layer. Input data are provided in the three-dimensional format: number of samples, time steps (24), and features (20).}
\label{tab:topology} 
\resizebox{0.18\textwidth}{!}{
\begin{tabular}{@{}cc@{}}
\toprule
\textbf{Layer} & \textbf{Output shape} \\ \midrule
Input          & 24x20                 \\
Dense          & 24x20                 \\
LSTM           & 24x18                 \\
LSTM           & 24x18                 \\
Dense          & 24x20                 \\ \bottomrule
\end{tabular}}
\end{table}

\begin{figure*}[h!]
  \caption{LSTM Autoencoder reconstruction error (MAE) on each attack type test set, with green representing normal CAN bus traffic and red malicious CAN bus traffic.}
  \centering
    \includegraphics[width=1.7\columnwidth]{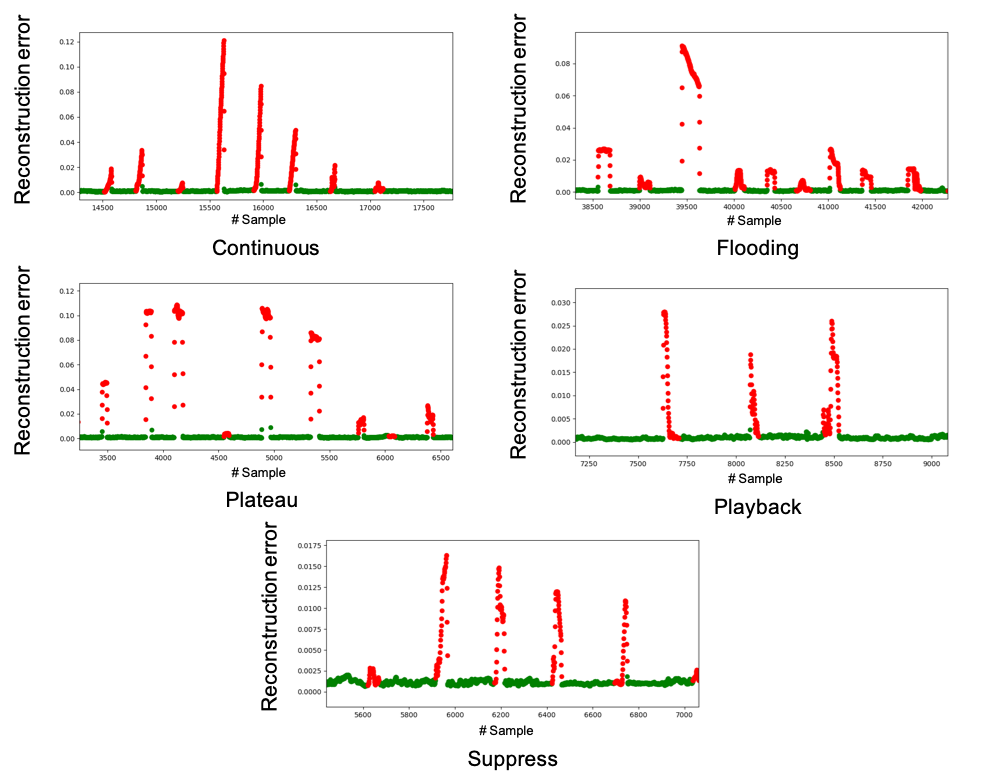}
    \label{fig:mae_anomaly}
\end{figure*}

The network hyperparameters were tuned using Keras-Tuner \cite{kerastun}. The anomaly score was evaluated using the Mean Absolute Error (MAE) between the true network traffic and its reconstruction, made by the Autoencoder. Figure \ref{fig:mae_anomaly} shows the reconstruction error (MAE) obtained on each attack type test set, with green representing normal CAN bus traffic and red malicious one. For all the attack types, the mean precision and recall are 0.86 and 0.81, respectively.

\section{Case Study 2: Capacity estimation in Lithium-Ion rechargeable batteries \label{sec:cas2}}

In this Section, we present a Convolutional Neural Network (CNN) LSTM architecture to predict the maximum releasable capacity of Li-Ion batteries, using the datasets made available by NASA \cite{nasa_db}. After embedding and validating the pre-trained NN into the aforementioned automotive MCUs, its complexity was profiled. 

\subsection{Dataset}

The data were extracted from one of the Li-Ion battery datasets made available by NASA Ames Prognostics Center of Excellence (PCoE) database \cite{nasa_db}. In all the experiments, Li-Ion rechargeable batteries were run through impedance, charge, and discharge operational profiles, measuring battery impedance, temperature, voltage, current, and capacity. Due to the greater variability of discharge experimental conditions, which implies a greater complexity of the estimation, in the analysis only discharge cycles have been considered. The battery capacity can be obtained starting from a fully charged battery and integrating the discharge current over time, until it reaches a certain threshold voltage \cite{Battery_manag}. Considering that the discharge may not be complete in real-world conditions, only some samples for each discharge cycle were selected. Output current, battery terminal voltage, temperature, and the time difference between samples were selected as model features. The capacity value corresponding to each discharge cycle was used as target for the prediction. Different batteries were used for the testing and training phases, thus getting closer to a real use case. During the training phase, a validation set was used to evaluate the loss function and to tune the parameters (hold-out method) \cite{bishop1995neural}.

\subsection{Convolutional and Long Short-Term Memory Network}

The architecture used for estimating the maximum releasable capacity was a CNN LSTM. In fact, the CNN structure allows the extraction of significant patterns from time series by reducing noise \cite{borovykh_conditional_2017}, and its temporal and spatial structure is particularly suitable for learning complex input features \cite{cnn_info}. Among Recurrent Neural Networks (RNNs), the LSTM architecture has been very successful with the long-term dependencies of time series \cite{Sutsk, lstmn}. Moreover, while standard RNNs experience the vanishing gradient problem, LSTM networks can overcome it. 

\begin{figure}[h!]
  \caption{The implemented CNN-LSTM Architecture consists of a 1D convolutional layer, a max pooling layer, a LSTM layer, and a dense output layer. Input data are provided in the three-dimensional format: number of samples, time steps (20), and features (4).}
  \centering
    \includegraphics[width=0.9\columnwidth]{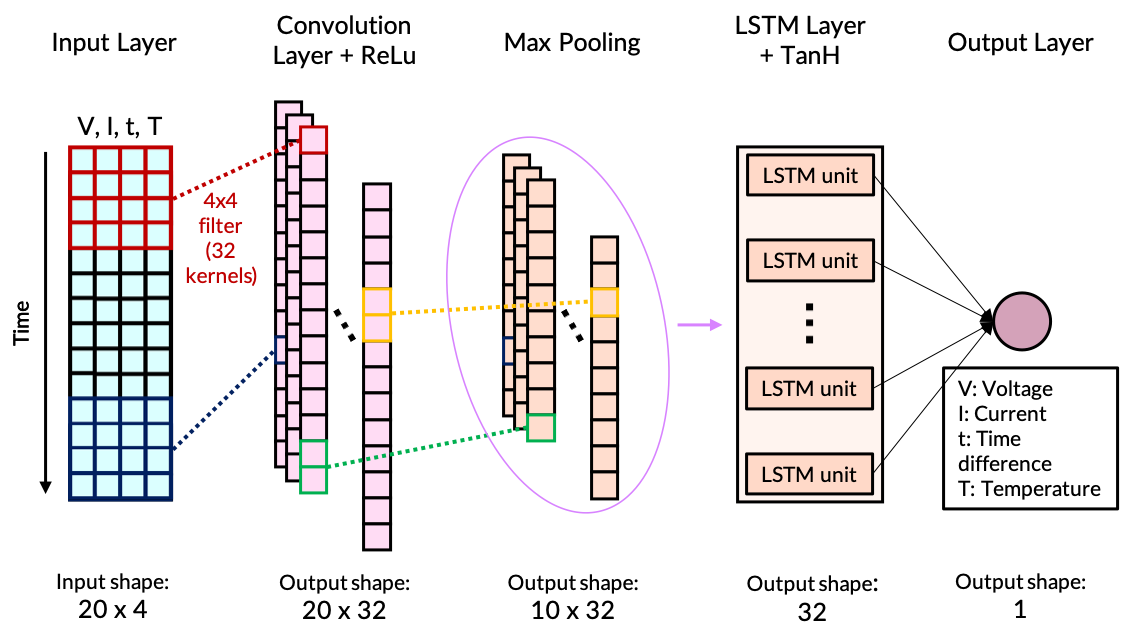}
    \label{fig:CNN_LSTM}
\end{figure}

The input data are provided in the format that CNN expects, i.e. the three-dimensional one: number of samples, time steps (20), and features (4). The convolutional layer is initialized with 32 filters, of size 4x4, and it uses the ReLu activation function after output normalization. A max pooling layer is added to summarize the input features. Then, an LSTM layer with 32 output units \cite{lstm_units} and with TanH activation function is used. The output is given by the dense single node output layer. The network is made by 8961 parameters. The developed CNN LSTM architecture is shown in Figure \ref{fig:CNN_LSTM}.

\begin{figure}[h!]
\centering
  \caption{Capacity estimation results for CNN LSTM versus ground truth values.}
  \centering
    \hspace*{-0.1cm}\includegraphics[width=0.9\columnwidth]{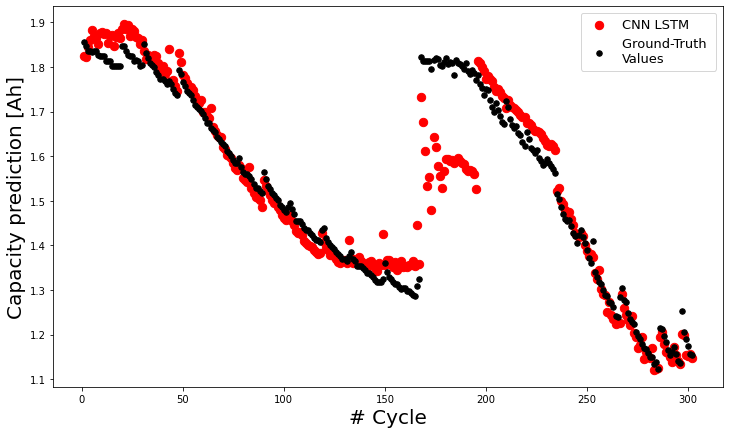}
    \label{fig:capp}
    \vspace{-2.5mm}
\end{figure}

Adaptive Moment Estimation (Adam) optimization algorithm was used to train the network, and the Mean Squared Error (MSE) was the chosen loss function to minimize. Features were scaled using MinMaxScaler, which preserves the shape of the original distribution \cite{minmax}. The capacity estimation error is computed using the MAE. Due to the randomness present during the training procedure (e.g. random weights initialization in NNs), at each run the results can be different. Thus, the model was trained and tested 10 times, each time using a different value for the pseudo-random number generator. The mean MAE obtained is 0.0434, below the EVs acceptable SOH error range of ${\pm}$0.05 \cite{indrnn}. The capacity estimation results of the CNN LSTM together with the ground truth values are shown in Figure \ref{fig:capp}. Further analysis has been made on the aforementioned dataset comparing different ML models \cite{9133084}, but it is out of the scope of this paper.

\section{Complexity Profiling}
\label{sec:compl}

The proposed NNs were evaluated with three automotive-grade MCUs (i.e., SPC584B, SPC58EC, and SPC58NH). The AI plug-in of the SPC5-STUDIO allowed the analysis of their performances in terms of Flash [Kb], Random Access Memory (RAM) [Kb], and average inference time [ms]. Tables \ref{table:lstmauto_compl} and \ref{table:cnnlstm_compl} show the results obtained, for each NN.

\begin{table}[h!]
\centering
\caption{Flash [Kb], RAM [Kb], and average inference time [ms] required by the LSTM Autoencoder, for each MCU.}
\label{table:lstmauto_compl} 
\resizebox{0.47\textwidth}{!}{
\begin{tabular}{@{}cccc@{}}
\toprule
\textbf{Device} & \textbf{Flash {[}Kb{]}} & \textbf{RAM {[}Kb{]}} & \textbf{Average inference time {[}ms{]}} \\ \midrule
SPC584B & 24.92 & 4.05 & 11 \\
SPC58EC & 24.92 & 4.05 & 8 \\
SPC58NH & 24.92 & 4.05 & 6 \\ \bottomrule
\end{tabular}}
\end{table}

\begin{table}[h!]
\centering
\caption{Flash [Kb], RAM [Kb], and average inference time [ms] required by the CNN LSTM, for each MCU.}
\label{table:cnnlstm_compl} 
\resizebox{0.47\textwidth}{!}{
\begin{tabular}{@{}cccc@{}}
\toprule
\textbf{Device} & \textbf{Flash {[}Kb{]}} & \textbf{RAM {[}Kb{]}} & \textbf{Average inference time {[}ms{]}} \\ \midrule
SPC584B & 35.13 & 2.25 & 6.34 \\
SPC58EC & 35.13 & 2.25 & 4.38 \\
SPC58NH & 35.13 & 2.25 & 3.86 \\ \bottomrule
\end{tabular}}
\end{table}

\begin{figure}[h!]
\centering
  \caption{Flash, RAM, and average execution time percentages obtained for each LSTM Autoencoder layer.}
  \centering
    \hspace*{-0.1cm}\includegraphics[width=0.9\columnwidth]{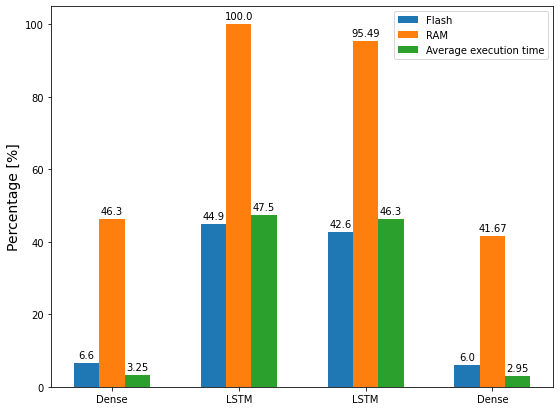}
    \label{fig:lstmauto_perc}
    \vspace{-2.5mm}
\end{figure}

\begin{figure}[h!]
\centering
  \caption{Flash, RAM, and average execution time percentages obtained for each CNN LSTM layer.}
  \centering
    \hspace*{-0.1cm}\includegraphics[width=0.9\columnwidth]{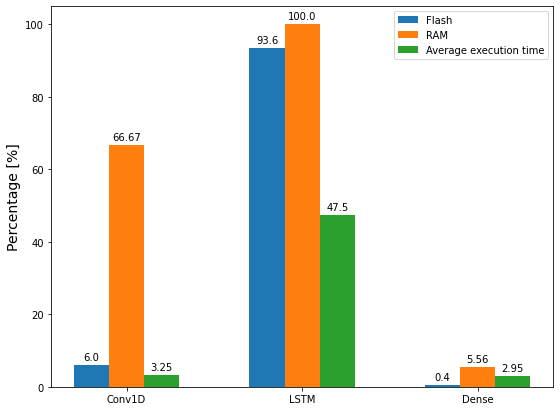}
    \label{fig:cnnlstm_perc}
    \vspace{-2.5mm}
\end{figure}

Note that only the average inference time differs between the MCUs, since it decreases linearly as the clock frequency increases. Flash, RAM, and the average run time percentages are shown in Figures \ref{fig:lstmauto_perc} and \ref{fig:cnnlstm_perc}, for each layer of each model. Due to its greater complexity, the LSTM layer is the most expensive both in terms of Flash (\%),  RAM (\%), and average execution time (\%), for both the architectures. The validation of the NNs was run on each of the chosen MCUs with 100\% cross-accuracy, which uses the outputs of the original model as ground truth values for those of the C-model.

\section{Conclusion\label{sec:con}}

In this work, we presented a framework for the implementation of Neural Network-based models on automotive-grade MCUs, applied to two applications of interest: intrusion detection on the CAN bus, and residual capacity estimation in Li-Ion batteries. The models training phases took place offline, and the pre-trained architectures were embedded on three automotive-grade MCUs, designed to deliver longevity, safety, and data integrity. The models were tested and validated, demonstrating their effectiveness, and their complexity on the MCUs was profiled.
Future works will focus on implementing the models taking into account the safety requirements of embedded automotive applications.
\bibliographystyle{ACM-Reference-Format}
\bibliography{acmart}


\end{document}